\documentclass[10pt,twocolumn,letterpaper]{article}

\usepackage{iccv}
\usepackage{times}
\usepackage{epsfig}
\usepackage{graphicx}
\usepackage{amssymb}
\usepackage{pifont}
\usepackage{amsmath}
\usepackage{amssymb}
\usepackage{dsfont}
\usepackage[ruled,vlined]{algorithm2e}
\renewcommand{\KwResult}{\textbf{Output:}}
\renewcommand{\KwData}{\textbf{Input:}}
\SetKwBlock{Repeat}{repeat}{}

\usepackage{makecell}
\usepackage{booktabs} 
\usepackage{array}  %
\usepackage{multirow}
\usepackage[tight,normalsize,sf,SF]{subfigure}
\usepackage{colortbl}
\usepackage{color}

\def\I{\mathbf{I}}
\def\Y{\mathbf{Y}}
\def\S{\mathbf{S}}
\usepackage{changes}

\definecolor{orange}{rgb}{1.0, 0.5, 0.0}

\usepackage[pagebackref=true,breaklinks=true,letterpaper=true,colorlinks,bookmarks=false]{hyperref}
\usepackage{breakurl}

\iccvfinalcopy 

\ificcvfinal\fi
\begin{document}

\title{Prior-aware Neural Network for Partially-Supervised \\ Multi-Organ Segmentation}

\author{Yuyin Zhou$^{1}$\thanks{This work was partly done when Yuyin Zhou, Chong Wang, Xinlei Chen and Mei Han were at Google.} \thanks{Equal Contribution.} \qquad
Zhe Li$^{2\dag}$ \qquad
Song Bai$^{3}$ \qquad
Chong Wang$^{4*}$ \qquad
Xinlei Chen$^{5*}$ \\
Mei Han$^{6*}$ \qquad
Elliot Fishman$^{7}$ \qquad
Alan L. Yuille$^1$ \vspace{.5em}\\
$^1$Johns Hopkins University \quad
$^2$Google AI \quad
$^3$University of Oxford \quad 
$^4$ByteDance Inc. \\
$^5$Facebook \quad
$^6$PAII Inc. \quad
$^7$The Johns Hopkins Medical Institute
\vspace{-1em}
}
\maketitle

\thispagestyle{empty}
\begin{abstract}
Accurate multi-organ abdominal CT segmentation is essential to many clinical applications such as computer-aided intervention. 
As data annotation requires massive human labor from experienced radiologists, it is common that training data are partially labeled,~\eg,~pancreas datasets only have the pancreas labeled while leaving the rest marked as background. However, these background labels can be misleading in multi-organ segmentation since the ``background'' usually contains some other organs of interest. 
To address the background ambiguity in these partially-labeled datasets, we propose Prior-aware Neural Network (PaNN) via explicitly incorporating anatomical priors on abdominal organ sizes, guiding the training process with domain-specific knowledge.
More specifically, PaNN assumes that the average organ size distributions in the abdomen should approximate their empirical distributions, prior statistics obtained from the fully-labeled dataset. 
As our training objective is difficult to be directly optimized using stochastic gradient descent,
we propose to reformulate it in a min-max form and optimize it via the stochastic primal-dual gradient algorithm. PaNN achieves state-of-the-art performance on the MICCAI2015 challenge ``Multi-Atlas Labeling Beyond the Cranial Vault'', a competition on organ segmentation in the abdomen. We report an average Dice score of $\textbf{84.97}\%$, surpassing the prior art by a large margin of $\textbf{3.27}\%$.
\end{abstract}

\section{Introduction}  \label{sec:intro}
This work focuses on multi-organ segmentation in abdominal regions which contain multiple organs such as liver, pancreas and kidneys. The segmentation of internal structures on medical images,~\eg,~CT scans, is an essential prerequisite for many clinical applications such as computer-aided diagnosis, computer-aided intervention and radiation therapy. Compared with other internal structures such as heart or brain, abdominal organs are much more difficult to segment due to the morphological and structural complexity, low contrast of soft tissues,~\etc 

\begin{figure}[tb]
\centering
\includegraphics[width=0.9\columnwidth]{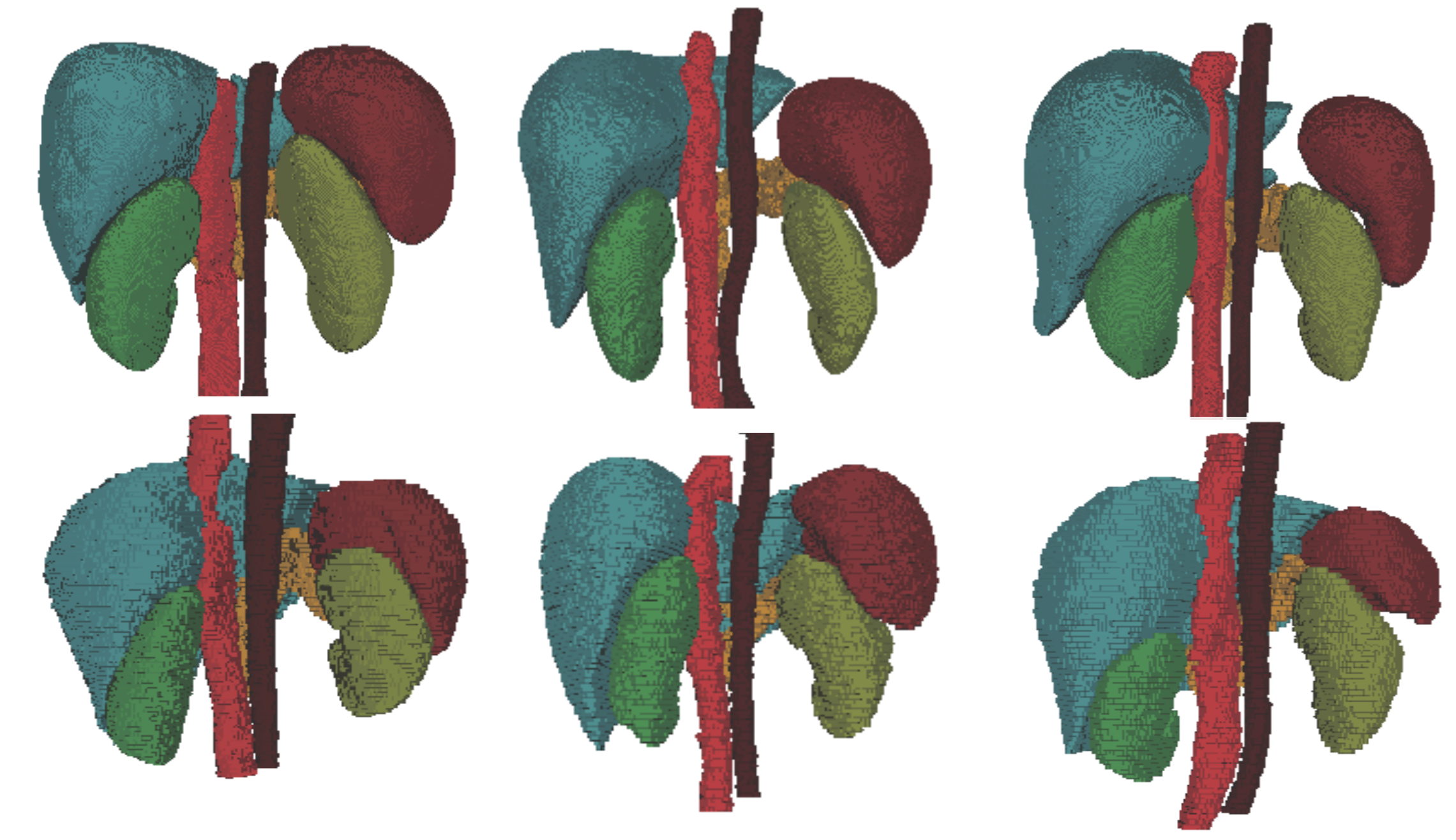}
\caption{3D Visualization of several abdominal organs (liver, spleen, left kidney, right kidney, aorta, inferior vena cava) to show the similarity of patient-wise abdominal organ size distributions.}
\label{fig:distribution}
\vspace{-1.5em}
 \end{figure}

\begin{figure*}
\centering
\includegraphics[width=2\columnwidth]{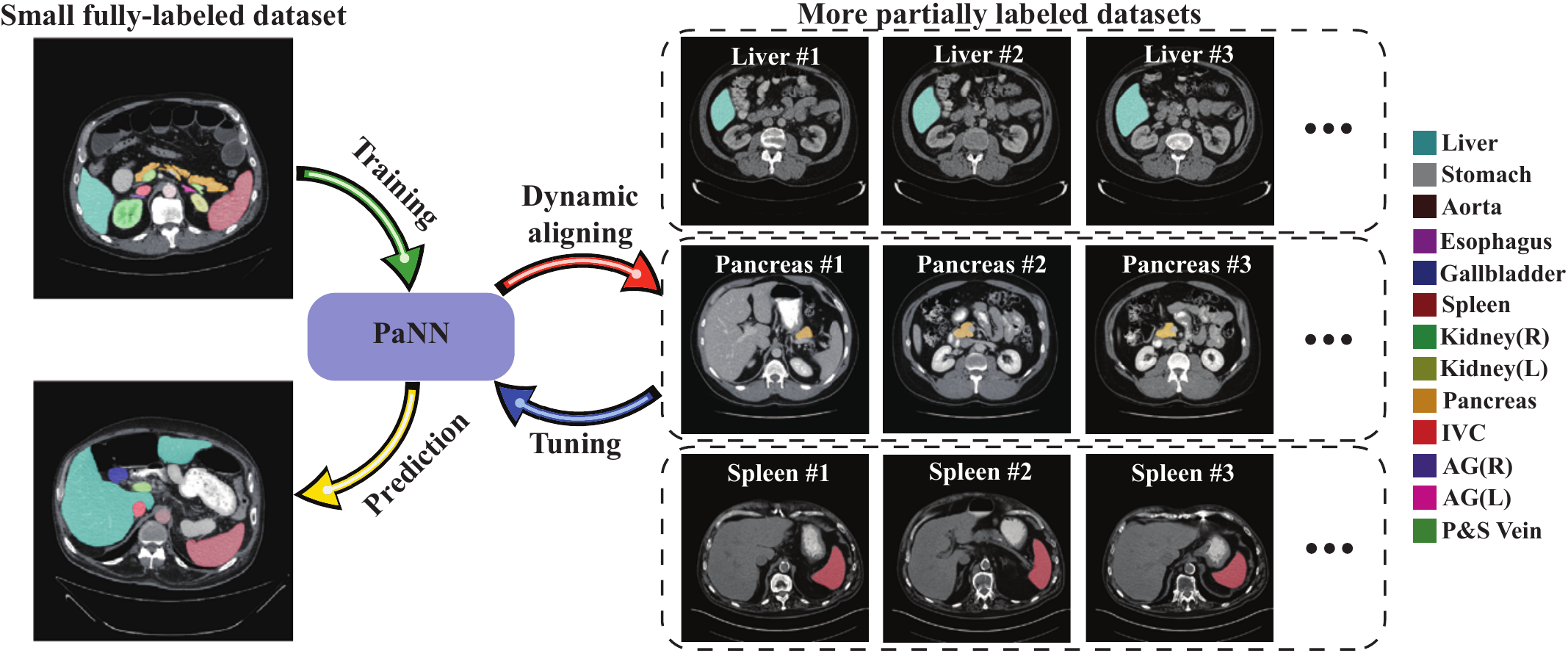}	
\caption{Overview of the proposed PaNN for partially-supervised multi-organ segmentation. It is trained with a small set of fully-labeled dataset and several partially-labeled datasets. The PaNN regularizes that the organ size distributions of the network output should approximate their prior statistics in the abdominal region obtained from the fully-labeled dataset.}
\label{fig:overview}
\vspace{-1em}
\end{figure*}

With the development of deep convolutional neural networks (CNNs), many medical image segmentation problems have achieved satisfactory results only when full-supervision is available~\cite{roth2018multi,roth2018spatial,zhou2017fixed,yu2018recurrent,ronneberger2015u,chen2017voxresnet}. Despite the recent progress, the annotation of medical radiology images is extremely expensive, as it must be handled by experienced radiologists and carefully checked by additional experts. This results in the lack of high-quality labeled training data. More critically, how to efficiently incorporate domain-specific expertise (\eg, anatomical priors) with segmentation models~\cite{dalca2018anatomical,oktay2018anatomically}, such as organ shape, size, remains an open issue.


Our key observation is that, in medical image analysis domain, instead of scribbles~\cite{lin2016scribblesup,tang2018normalized,tang2018regularized} , points~\cite{bearman2016s} and image-level tags~\cite{papandreou2015weakly,pathak2014fully,xu2014tell}, there exists a considerable number of datasets in the form of abdominal CT scans~\cite{roth2015deeporgan, roth2018multi, simpson2019large}. To meet different research goals or practical usages, these datasets are annotated to target different organs (a subset of abdominal organs), \eg, pancreas datasets~\cite{roth2015deeporgan} only have the pancreas labeled while leaving the rest marked as background. 

The aim of this work is to fully leverage these existing partially-annotated datasets to assist multi-organ segmentation, which we refer to as \emph{partial supervision}. 
To address the challenge of partial supervision, an intuitive solution is to simply train a segmentation model directly on both the labeled data and the partially-labeled data in the semi-supervised manner~\cite{rajchl2017deepcut, bai2017semi, papandreou2015weakly}. However, it 1) fails to take advantages of the fact that medical images are naturally more constrained compared with natural images~\cite{nosrati2016incorporating}; 2) is intuitively misleading as it treats the unlabeled pixels/voxels as background. To overcome these issues, we propose Prior-aware Neural Network (PaNN) to handle such background ambiguity via incorporating prior knowledge on organ size distributions. We achieve this via a prior-aware loss, which acts as an auxiliary and soft constraint to regularize that the average output size distributions of different organs should approximate their prior proportions.  Based on the anatomical similarities (Fig.~\ref{fig:distribution}) across different patient scans~\cite{dalca2018anatomical, oktay2018anatomically, kervadec2019constrained}, the prior proportions are estimated by statistics from the fully-labeled data. The overall pipeline is illustrated in Fig.~\ref{fig:overview}.
It is important to note that the training objective is hard to be directly optimized using stochastic gradient descent.
To address this issue, we propose to formulate our objective in a min-max form, which can be well optimized via the stochastic primal-dual gradient algorithm~\cite{liu2017unsupervised}. To summarize, our contributions are three-fold:

\noindent\textbf{1)}~We propose Prior-aware Neural Network, which incorporates domain-specific knowledge from medical images, to facilitate multi-organ segmentation via using partially-annotated datasets.

\noindent\textbf{2)}~As the training objective is difficult to be directly optimized using stochastic gradient descent, it is essential to reformulate it in a min-max form and optimize via stochastic primal-dual gradient~\cite{liu2017unsupervised}.

\noindent\textbf{3)}~PaNN significantly outperforms previous state-of-the-arts even using fewer annotations. It achieves $84.97\%$ on the MICCAI2015 challenge ``Multi-Atlas Labeling Beyond the Cranial Vault'' in the free competition for organ segmentation in the abdomen.

\section{Related Work} \label{sec:related}
Currently, the most successful deep learning techniques for semantic segmentation stem from a common forerunner,~\ie,~Fully Convolutional Network (FCN)~\cite{long2014fully}. Based on FCN, many recent advanced techniques have been proposed, such as DeepLab~\cite{chen2015semantic, chen2018deeplab, chen2018encoder}, SegNet~\cite{badrinarayanan2017segnet}, PSPNet~\cite{zhao2017pyramid}, RefineNet~\cite{lin2017refinenet},~\etc Most of these methods are based on supervised learning, hence requiring a sufficient number of labeled training data to train. 
To cope with scenarios where supervision is limited, researchers begin to investigate the weakly-supervised setting~\cite{papandreou2015weakly,pathak2014fully,dai2015boxsup},~\eg,~only bounding-boxes or image-level labels are available, and the semi-supervised setting~\cite{papandreou2015weakly, souly2017semi},~\ie,~unlabeled data are used to enlarge the training set. Papandreou~\etal~\cite{papandreou2015weakly}~propose EM-Adapt where the pseudo-labels of the unknown pixels are estimated in the expectation step and standard SGD is performed in the maximization step. Souly \etal~\cite{souly2017semi} demonstrate the usefulness of generative adversarial networks for semi-supervised segmentation.

In the medical imaging domain, it becomes more intractable to acquire sufficient labeled data due to the difficulty of annotation, as the annotation has to be done by experts. Although fully-supervised methods (\eg,~UNet~\cite{ronneberger2015u}, VoxResNet~\cite{chen2017voxresnet}, DeepMedic~\cite{kamnitsas2017efficient}, 3D-DSN~\cite{dou20163d}, HNN~\cite{roth2018spatial}) have achieved remarkable performance improvement in tasks such as brain MR segmentation, abdominal single-organ segmentation and multi-organ segmentation, semi- or weakly-supervised learning is still a far more realistic solution. For example, Bai~\etal~\cite{bai2017semi} proposed an EM-based iterative method, where a CNN is alternately trained on labeled and post-processed unlabeled sets. In~\cite{zhang2017deep}, supervised and unsupervised adversarial costs are involved to address semi-supervised gland segmentation. DeepCut~\cite{rajchl2017deepcut} shows that weak annotations such as bounding-boxes in medical image segmentation can also be utilized by performing an iterative optimization scheme like~\cite{papandreou2015weakly}.

However, these methods fail to capture the anatomical priors~\cite{litjens2017survey}. Inclusion of priors in medical imaging could potentially have much more impact compared with their usage in natural images since anatomical objects in medical images are naturally more constrained in terms of shape, location, size, \etc. Some recent works~\cite{dalca2018anatomical, oktay2018anatomically} demonstrate that these priors can be learned by a generative model. But these methods will induce heavy computational overhead. Kervadec \etal~\cite{kervadec2019constrained} proposed that  directly imposing inequality constraints on sizes is also an effective way of incorporating anatomical priors. Unlike these methods, we propose to learn from partial annotations by embedding the abdominal region statistics in the training objective, which requires no additional training budget.

\section{Prior-aware Neural Network}
Our work aims to address the multi-organ segmentation problem with the help of multiple existing partially-labeled datasets. Given a CT scan where each element indicates the Housefield Unit (HU) of a voxel, the goal is to find the predicted labelmap of each pixel/voxel. 

\subsection{Partial Supervision} \label{sec:partial}
We consider a new supervision paradigm,~\ie,~partial supervision, for multi-organ segmentation. This is motivated by the fact that there exists a considerable number of datasets with only one or a few organs labeled  in the form of abdominal CT scans~\cite{roth2015deeporgan, roth2018multi, simpson2019large} in medical image analysis, which can serve as partial supervision for multi-organ segmentation (see the list in the appendix). Based on domain knowledge, our approach assumes the following characteristics of the datasets which are common in medical image analysis. First, the scanning protocols of medical images are well standardized, \eg, brain, head and neck, chest, abdomen, and pelvis in CT scans, which means that the internal structures are consistent in a limited range according to the scanning protocol (see Fig.~\ref{fig:distribution}). Second, internal organs have anatomical and spatial relationships such as gastrointestinal track, \ie, stomach, duodenum, small intestine, and colon are connected in a fixed order. 

The partially-supervised setting can be formally defined as below. Given a fully-labeled dataset $\S_\textup{L}=\{\I_\textup{L},\Y_\textup{L}\}$ with the annotation $\Y_\textup{L}$ known and T partially-labeled datasets $\S_\textup{P}=\{\S_{\textup{P}_1},\S_{\textup{P}_2},...\S_{\textup{P}_T}$\} with the $t$-th dataset defined as $\S_{\textup{P}_t}=\{\I_{\textup{P}_t},\Y_{\textup{P}_t}\}$. $\textup{L} = \{1,2,...,n_\textup{L}\}$ and $\textup{P}_t = \{1,2,...,n_{\textup{P}_t}\}$ denote the image indices for $\S_\textup{L}$ and  $\S_{\textup{P}_t}$, respectively.
For each element $y_{ij} \in \Y_{\textup{L}}$, $y_{ij}$ denotes the annotation of the $j$-th pixel in the $i$-th image $\I_i\in\I_{\textup{P}_t}$ and is selected from $\mathcal{L}$, where $\mathcal{L}$ denotes the abdominal organ space,~\ie,~$\mathcal{L}=\{\text{spleen},\text{pancreas},\text{liver},...\}$. For the $t$-th partially-labeled dataset $\S_{\textup{P}_t}$, $y_{ij}\in\Y_{\textup{P}_t}$ is selected from $\mathcal{L}_{\textup{P}_t} \subseteq \mathcal{L}$. In 2D-based segmentation models, the $i$-th input $\I_i$ is a sliced 2D image from either Axial, Coronal or Saggital view of the whole CT scan~\cite{zhou2017fixed,roth2018spatial,zhou2019semi,wang2018training}. In 3D-based segmentation models, $\I_i$ is a cropped 3D patch from the whole CT volume~\cite{cciccek20163d, milletari2016v}. 
Note that semi-supervision and fully-supervision are two extreme cases of partial supervision, when the set of partial labels is an empty set ($\mathcal{L}_{\textup{P}_t} = \oslash$) and is equal to the complete set ($\mathcal{L}_{\textup{P}_t} = \mathcal{L}$), respectively.

A naive solution is to simply train a segmentation network from both the fully-labeled data and the partially-labeled data and alternately update the network parameters and the segmentations (pseudo-labels) for the partially-labeled data~\cite{zhou2019semi,bai2017semi}. While these EM-like approaches have achieved significant improvement compared with fully-supervised methods, they require high-quality pseudo-labels and fail to explicitly incorporate anatomical priors on shape or size.

To address this issue, we propose a Prior-aware Neural Network (PaNN), aiming at explicitly embedding anatomical priors without incurring any additional budget. More specifically, the anatomical priors are enforced by introducing an additional penalty which acts as a soft constraint to regularize that the average output distributions of organ sizes should mimic their empirical proportions. This prior is obtained by calculating the organ size statistics of the fully-labeled dataset. An overview of the overall framework is shown in Fig.~\ref{fig:overview}, and the detailed training procedures will be introduced in the following sections.

\subsection{Prior-aware Loss}

Consider a segmentation network parameterized by $\boldsymbol{\Theta}$, which outputs probabilities $\mathbf{p}$.
Let $\mathbf{q}\in\mathbb{R}^{(|\mathcal{L}| + 1)\times 1}$ be the label distribution in the fully-labeled dataset, with $q^l$ describing the proportion of the $l$-th label (organ). Then, we estimate the average predicted distribution of the pixels in the partially-labeled datasets as
\begin{equation} \label{eq:bar_p}\small
\vspace{-0.5em}
\mathbf{\bar{p}}=\frac1N\sum_{t=1}^T\sum_{i\in\textup{P}_t}\sum_j{\mathbf{p}_{ij}},
\end{equation}
where $\mathbf{p}_{ij} = [p_{ij}^{0}, p_{ij}^{1}, ..., p_{ij}^{|\mathcal{L}|}]$ denotes
the probability vector of the $j$-th pixel in the $i$-th input slice $\I_i$, and $N$ is the total number of pixels/voxels. Recall that $T$ is the total number of partially-labeled datasets.

To embed the prior knowledge, the prior-aware loss is defined as 
\begin{equation}\label{eqn:klloss}\small
\begin{aligned} 
&\textstyle {\rm KL}_{\rm marginal} (\mathbf{q}|\mathbf{\bar{p}})   \triangleq \sum_l {\rm KL} (q^l | \bar{p}^l) \\  
&\textstyle = -\sum_l\left(  q^l \log \bar{p}^l + (1-q^l) \log( 1- \bar{p}^l)\right) + \emph{const} \\
&\textstyle = -\{\mathbf{q}\log\mathbf{\bar{p}} + (1 - \mathbf{q})\log(1 - \mathbf{\bar{p}})\} + \emph{const},
\end{aligned}
\end{equation}

\noindent{which measures the matching probability of the two distributions $\mathbf{q}$ and $\mathbf{\bar{p}}$ via Kullback-Leibler divergence. Note that each class is treated as one vs. rest when calculating the matching probabilities.
Therein, the rationale of Eq.~\eqref{eqn:klloss} is that the output distributions $\mathbf{\bar{p}}$ of different organ sizes should approximate their empirical marginal proportions $\mathbf{q}$, which generally reflects the domain-specific knowledge.}

Note that $\mathbf{q}$ is a global estimation of label distribution of the fully-labeled training data, which remains unchanged. Consequently, $\rm{H}(\mathbf{q})$ is constant which can be omitted during the network training. Nevertheless, we observe that it is still problematic to directly apply stochastic gradient descent, as we will detail in Sec.~\ref{sec:derivation}.

Specifically in our case, our final training objective is
\begin{equation}  \label{eq:overall_loss}\small
\min_{\boldsymbol{\Theta}, \Y_\textup{P}}\mathcal{J}_\textup{L}(\boldsymbol{\Theta}) + \lambda_1\mathcal{J}_\textup{P}(\boldsymbol{\Theta, \Y_\textup{P}})+\lambda_2\mathcal{J}_\textup{C}(\boldsymbol{\Theta}),
\end{equation}
where $\mathcal{J}_\textup{L}(\boldsymbol{\Theta})$ and $\mathcal{J}_\textup{P}(\boldsymbol{\Theta, \Y_\textup{P}})$ are the cross entropy loss on the fully-labeled data and the partially-labeled data, respectively. And $\Y_\textup{P}$ denotes the computed pseudo-labels as well as existing partial labels
from the partially-labeled dataset(s). Note that the prior-aware loss $\mathcal{J}_\textup{C}$ is used as a soft global constraint to stablize the training process. Concretely, $\mathcal{J}_\textup{L}(\boldsymbol{\Theta})$ is defined as
\begin{equation} \label{eq:cross_fully}\small
\mathcal{J}_\textup{L}=-\frac1N\sum_{i\in\textup{L}}\sum_{j}\sum_{l=0}^{|\mathcal{L}|}{\mathds{1}(y_{ij}=l)\log p_{ij}^l},
\end{equation}
where $p_{ij}^l$ denotes the softmax probability of the $j$-th pixel in the $i$-th image to the $l$-th category. $\mathcal{J}_\textup{P}(\boldsymbol{\Theta, \Y_\textup{P}})$ is given by
\begin{equation} \label{eq:loss_partial}
\begin{aligned}
  \mathcal{J}_\textup{P}=-\frac1N\sum_{t=1}^{T}\sum_{i\in\textup{P}_t}\sum_{j}\sum_{l=0}^{|\mathcal{L}|}
  \{&\mathds{1}(y_{ij}=l)\log p_{ij}^l\\
  + &\mathds{1}(y'_{ij}=l)\log p_{ij}^l\}, 
\end{aligned}
\end{equation}
where the first term corresponds to the pixels with their labels $\Y_\textup{P}$ given, \ie, $y_{ij} \in \mathcal{L}_{\textup{P}_t}$. The second term corresponds to unlabeled background pixels, and $\Y_\textup{P}$ needs to be estimated during the model training as a kind of pseudo-supervision, \ie, $y'_{ij} \in \mathcal{L} - \mathcal{L}_{\textup{P}_t}$.

\subsection{Derivation} \label{sec:derivation}

By substituting Eq.~(\ref{eq:bar_p}) into Eq.~(\ref{eqn:klloss}) and expanding $\mathbf{q}, \mathbf{\bar{p}}$ into scalars,
we rewrite Eq.~(\ref{eqn:klloss}) as 
\begin{equation}
\label{eq:klloss_new}
    \begin{aligned}
    \mathcal{J}_\textup{C} 
    =-\sum_{l=0}^{|\mathcal{L}|}\{{q^l}&\log{\frac1N\sum_{t=1}^T\sum_{i \in \textup{P}_t}\sum_jp_{ij}^l} + \\
   (1 - q^l)&\log(1 - {\frac1N\sum_{t=1}^T\sum_{i \in \textup{P}_t}\sum_jp_{ij}^l})\} + const.
    \end{aligned}
\end{equation}
From Eq.~\eqref{eqn:klloss} and Eq.~\eqref{eq:klloss_new} we can see that the average distribution $\mathbf{\bar{p}}$ of organ sizes is inside the logarithmic loss,
which is very different from standard machine learning loss such as Eq.~\eqref{eq:cross_fully} and Eq.~\eqref{eq:loss_partial} where the
average is outside logarithmic loss. And directly minimizing by stochastic gradient descent is very difficult as the true gradient induced by Eq.~\eqref{eqn:klloss} is not a summation of independent terms, the stochastic gradients would be intrinsically biased~\cite{liu2017unsupervised}.

To remedy this, we propose to optimize the KL divergence term using stochastic primal-dual gradient~\cite{liu2017unsupervised}. Our goal here is to transform the prior-aware loss into an equivalent min-max problem by taking the sample average out of the logarithmic loss. 
We introduce two auxiliary variables to assist the optimization,~\ie,~the primal variable $\alpha$ and the dual variable $\beta$. First, the following identity holds 
\begin{equation} \label{eq:primal_dual}\small
-\log\alpha=\max_\beta\left(\alpha\beta+1+\log(-\beta)\right)
\end{equation} due to the property of the log function. Based on Eq.~\eqref{eq:primal_dual}, we define $\boldsymbol{\nu}\in\mathbb{R}^{|\mathcal{L}|\times 1}$ as the dual variable associated to the primal variable $\mathbf{\bar{p}}$, and define $\boldsymbol{\mu}\in\mathbb{R}^{|\mathcal{L}|\times 1}$ as the dual variable associated to the primal variable $(1-\mathbf{\bar{p}})$. Then, we have
\begin{equation} \label{eq:primal_dual1}\small
\begin{split}
-\log\bar{p}^l&=\max_{\nu^l}\left(\bar{p}^l\nu^l+1+\log(-\nu^l)\right) \\
-\log(1-\bar{p}^l)&=\max_{\mu^l}\left((1-\bar{p}^l)\mu^l+1+\log(-\mu^l)\right),
\end{split}
\end{equation}
where $\nu^l$ (or $\mu^l$) denotes the $l$-th element of $\boldsymbol{\nu}$ (or $\boldsymbol{\mu}$). Substituting them into Eq.~\eqref{eqn:klloss}/Eq.~\eqref{eq:klloss_new}, maximizing the KL divergence is equivalent to the following min-max optimization problem:

\begin{equation}\label{eqn:overall_optimization} \small
\begin{split}
\min_{\boldsymbol{\Theta}}\max_{\boldsymbol{\nu},\boldsymbol{\mu}}&~\sum_{l}q^l\left(\bar{p}^l\nu^l+1+\log(-\nu^l)\right)\\
+&~\sum_{l}(1-q^l)\left((1-\bar{p}^l)\mu^l+1+\log(-\mu^l)\right) \\
\Leftrightarrow\min_{\boldsymbol{\Theta}}\max_{\boldsymbol{\nu},\boldsymbol{\mu}}&~\sum_{l}\left(q^l{\nu^l}-(1-q^l){\mu^l}\right)\bar{p}^{l}+q^l\log(-{\nu^l}) \\
+&~\sum_{l}(1-q^l)\left(\mu^l+\log(-\mu^l)\right),
\end{split}
\end{equation}
which brings the sample average out of the logarithmic loss. Note that we ignore the constant in the above formulas.

\begin{algorithm}[t!]\small
\DontPrintSemicolon
\KwData{\\Fully-labeled training data $\S_\textup{L}$;
\\Partially-labeled training data $\S_\textup{P}$;
\\ Hyperparameters: $\lambda_1$, $\lambda_2$;
\\}
\KwResult{\\Segmentation model $\boldsymbol{\Theta}$;
\\ }
\Begin{
Train the segmentation model $\boldsymbol{\Theta}$ on $\S_\textup{L}$; \\
Compute the prior distribution $\mathbf{q}$ on $\S_\textup{L}$; \\
Initialize $\boldsymbol{\nu}=-1/\mathbf{q}$ and $\boldsymbol{\mu}=1/(1-\mathbf{q})$; \\
\Repeat
{
  Estimate pesudo-labels $\Y_\textup{P}$ with $\boldsymbol{\Theta}$; \\
  Update $\boldsymbol{\nu}$ and $\boldsymbol{\mu}$ via stochastic gradient ascent; \\
  Update $\boldsymbol{\Theta}$ via stochastic gradient descent; \\
}
}
\KwRet{$\boldsymbol{\Theta}$}
\caption{\small The training procedure of PaNN}
\label{Algo:minmax}
\end{algorithm}

\subsection{Model Training}
\label{sec:model_training}
We consider training a fully convolutional network~\cite{long2014fully, chen2018deeplab, ronneberger2015u} for multi-organ segmentation, where the input images are either 2D slices~\cite{wang2018training, roth2018spatial,zhou2017fixed} or 3D cropped patches~\cite{cciccek20163d, milletari2016v}. The training procedure can be divided into two stages.

In the first stage, we only train on the fully-labeled dataset $\S_\textup{L}$ by optimizing Eq.~\eqref{eq:cross_fully} via stochastic gradient descent (also means $\lambda_1=0$ and $\lambda_2=0$ in Eq.~\eqref{eq:overall_loss}). The goal of this stage is to find a proper initialization $\boldsymbol{\Theta_0}$ for the network weights, which stabilizes the later training procedure. 

In the second stage, we train the model on the union of the fully-labeled dataset $\mathbf{S}_\textup{L}$ and partially-labeled dataset(s) $\mathbf{S}_\textup{P}$ via Eq.~\eqref{eq:overall_loss}. As can be drawn, we have two groups of variables,~\ie,~the network weights $\boldsymbol{\Theta}$ and the three auxiliary variables $\{\boldsymbol{\nu},\boldsymbol{\mu},\Y_\textup{P}\}$. We adopt an alternating optimization, which can be decomposed into two subproblems:

    \vspace{1ex}\noindent\textbf{$\bullet$ Fixing $\boldsymbol{\Theta}$, Updating $\{\boldsymbol{\nu},\boldsymbol{\mu},\Y_\textup{P}\}$.}~With the network weights $\boldsymbol{\Theta}$ given, we can first estimate the pesudo-labels $\Y_\textup{P}$ of background pixels in the partially-labeled dataset(s) $\S_\textup{P}$. 
    Meanwhile, the optimization of $\boldsymbol\nu$ and $\boldsymbol\mu$ is a maximization problem. Hence, we do stochastic gradient {\it ascent} to learn $\boldsymbol\nu$ and $\boldsymbol\mu$. As for the initialization, we set $\boldsymbol{\nu}$ to $-1/\mathbf{q}$ and set $\boldsymbol{\mu}$ to $-1/(1-\mathbf{q})$, respectively.
    
    \vspace{1ex}\noindent\textbf{$\bullet$ Fixing $\{\boldsymbol{\nu},\boldsymbol{\mu},\Y_\textup{P}\}$, Updating $\boldsymbol{\Theta}$.}~By fixing the three auxiliary variables, we can then update the network weights $\boldsymbol{\Theta}$ via the standard stochastic gradient {\it descent}.

As can be seen, our algorithm is formulated as a min-max optimization. We summarize the detailed procedure of optimization in Algorithm~\ref{Algo:minmax}.

\section{Experiments}
\subsection{Experiment Setup}
\noindent\textbf{Datasets and Evaluation Metric.}~We use the training set released in the MICCAI 2015 Multi-Atlas Abdomen Labeling Challenge as the fully-labeled dataset $\S_\textup{L}$, which contains 30 abdominal CT scans with $3779$ axial contrast-enhanced abdominal clinical CT images in total. For each case, $13$ anatomical structures are annotated, including spleen, right kidney, left kidney, gallbladder, esophagus, liver, stomach, aorta, inferior vena cava (IVC), portal vein \& splenic vein, pancreas, left adrenal gland, right adrenal gland. Each CT volume consists of $85\sim198$ slices of $512\times512$ pixels, with a voxel spatial resolution of $([0.54\sim0.54]\times[0.98\sim0.98]\times[2.5\sim5.0])\textup{mm}^3$.

As for the partially-labeled dataset(s) $\S_\textup{P}$, we use a spleen segmentation dataset\footnote{Available at~\url{http://medicaldecathlon.com}} (referred as \textbf{A}), a pancreas segmentation dataset\footnote{Available at~\url{https://wiki.cancerimagingarchive.net/display/Public/Pancreas-CT}} (referred as \textbf{B}) and a liver segmentation dataset\textcolor{red}{$^1$} (referred as \textbf{C}). To make these partially-labeled datasets balanced, 40 cases are evenly selected from each dataset to constitute the partial supervision.

Following the standard cross-validation evaluation~\cite{roth2018multi, roth2018spatial, nogues2016automatic, zhou2017fixed, wang2018training}, we randomly partition the fully-labeled dataset $\S_L$ into $5$ complementary folds, each of which contains $6$ cases, then apply the standard 5-fold cross-validation. For each fold, we use $4$ folds (\ie,~$24$ cases) as full supervision and test on the remaining fold. 

The evaluation metric we use is the Dice-S{\o}rensen Coefficient (DSC), which measures the similarity between the prediction voxel set $\mathcal{Z}$ and the ground-truth set $\mathcal{Y}$. Its mathematical definition is  ${\mathrm{DSC}\!\left(\mathcal{Z},\mathcal{Y}\right)}={\frac{2\times\left|\mathcal{Z}\cap\mathcal{Y}\right|}{\left|\mathcal{Z}\right|+\left|\mathcal{Y}\right|}}$.
We report an average DSC of all the testing cases over the $13$ labeled anatomical structures for performance evaluation.

\vspace{1ex}\noindent\textbf{Implementation Details.}~Similar to~\cite{zhou2017fixed, roth2018spatial, roth2018multi, wang2018training}, we use the soft tissue CT window range of $[-125,275]$ HU. The intensities of each slice are then rescaled to $[0.0, 255.0]$. Random rotation of $[0,15]$ is used as an online data augmentation. Our implementations are based on the current state-of-the-art 2D\footnote{\url{https://github.com/tensorflow/models/tree/master/research/deeplab}}~\cite{chen2018encoder, chen2018deeplab} and 3D models\footnote{\url{https://github.com/DLTK/DLTK}}~\cite{ronneberger2015u,pawlowski2017dltk}. We provide an extensive study about how partially-labeled datasets facilitate multi-organ segmentation task and list thorough comparisons under different settings. 

As described in Sec.~\ref{sec:model_training}, the whole training procedure is divided into two stages. The first stage is the same as fully-supervised training, \ie, we train exclusively on the fully-labeled dataset $\mathcal{S}_\textup{L}$ for a certain number of iterations M$1$.
 
In the second stage, we switch to the min-max optimization on the union of the fully-labeled dataset and partially-labeled datasets for M$2$ iterations. 
In each mini-batch, the sampling rate of labeled data and partially-labeled data is $3:1$. It has been suggested~\cite{bai2017semi} that it is less necessary to update the pseudo-label $\mathbf{Y}_\textup{P}$ per iteration. Hence, $\mathbf{Y}_\textup{P}$ is updated every $10$K iterations in practice. In addition, the hyperparameters $\lambda_1$ and $\lambda_2$ are set to be $1.0$ and $0.1$, respectively. 
The same decay policy of learning rate is utilized as that used in the first stage. In the second stage, the initial learning rate for the minimization step and the maximization step are set as $10^{-5}$ and $2\times 10^{-5}$, respectively. 

For 2D implementations, the initial learning rate of the first stage is $2\times 10^{-5}$ and a \emph{poly} learning rate policy is employed. M$1$ and M$2$ are set as $40$K and $30$K, respectively. Following~\cite{roth2018multi, chen2018encoder, kamnitsas2017efficient}, we apply multi-scale inputs (scale factors are $\{0.75, 1.0, 1.25, 1.5, 1.75, 2.0\}$) in both training and testing phase. For 3D implementations, the initial learning rate of the first stage is $5e^{-4}$ and a fixed learning rate policy is employed. M$1$ and M$2$ are set as $80$K and $100$K, respectively.

\subsection{Experimental Comparison}

\newcommand{\cmark}{\ding{51}}
\begin{table}[tb]
    \centering
    \resizebox{0.95\linewidth}{!}{
    \begin{tabular}{l|l|l l l|c}
        \toprule[0.2em]
        \multirow{3}{*}{\textbf{Model}}& \multirow{3}{*}{\textbf{Supervision}} & \multicolumn{3}{c|}{\textbf{Partially-labeled}} & \multirow{3}{*}{\textbf{Average Dice}} \\
         & &   \multicolumn{3}{c|}{\textbf{dataset}}   & \\\cline{3-5}
         & & A \hspace{1em} & B \hspace{1em} & C &  \\
        \toprule[0.2em]
        \multirow{13}{*}{ResNet50~\cite{he2016deep}}& Full &  &  &  & 0.7535 \\
        \cline{2-6}
        & \multirow{4}{*}{Semi \cite{bai2017semi}} & \cmark &  &  & 0.7593  \\
        & &  & \cmark &  & 0.7632  \\
        & &  &  & \cmark & 0.7596  \\
        & & \cmark & \cmark & \cmark & \textbf{0.7669}  \\
        \cline{2-6}
        & \multirow{4}{*}{Partial \textbf{(ours)}} & \cmark &  &  & 0.7650  \\
        & &  & \cmark &  & 0.7662  \\
        & &  &  & \cmark & 0.7631  \\
        & & \cmark & \cmark & \cmark & \textbf{0.7705}  \\
        \cline{2-6}
        & \multirow{4}{*}{PaNN \textbf{(ours)}} & \cmark &  &  & 0.7716 \\
        & &  & \cmark &  & 0.7712  \\
        & &  &  & \cmark & 0.7705   \\
        & & \cmark & \cmark & \cmark & \underline{\textbf{0.7833}}  \\
        \midrule
        \midrule
        \multirow{13}{*}{ResNet101~\cite{he2016deep}}& Full &  &  &  & 0.7614 \\
        \cline{2-6}
        & \multirow{4}{*}{Semi \cite{bai2017semi}} & \cmark &  &  & 0.7637  \\
        & &  & \cmark &  & 0.7649  \\
        & &  &  & \cmark & 0.7647  \\
        & & \cmark & \cmark & \cmark & \textbf{0.7719} \\
        \cline{2-6}
        & \multirow{4}{*}{Partial \textbf{(ours)}} & \cmark &  &  & 0.7714 \\
        & &  & \cmark &  & 0.7695  \\
        & &  &  & \cmark & 0.7684    \\
        & & \cmark & \cmark & \cmark & \textbf{0.7735}  \\
        \cline{2-6}
        & \multirow{4}{*}{PaNN \textbf{(ours)}} & \cmark &  &  & 0.7770  \\
        & &  & \cmark &  & 0.7819 \\
        & &  &  & \cmark & 0.7748  \\
        & & \cmark & \cmark & \cmark & \underline{\textbf{0.7904}} \\
        \midrule
        \midrule
        \multirow{4}{*}{3D-UNet~\cite{cciccek20163d}} & 3D-UNet-fully-sup &  &  &  & 0.7066 \\
        \cline{2-6}
        & Semi \cite{bai2017semi} & \cmark &\cmark  &\cmark & 0.7193  \\
        & Partial \textbf{(ours)}&\cmark & \cmark &\cmark & 0.7163  \\
        & PaNN \textbf{(ours)}&\cmark &\cmark & \cmark & \underline{\textbf{0.7208}}  \\
        \bottomrule[0.2em]
    \end{tabular}
    }
    \caption{Performance comparison (DSC) with fully-supervised and semi-supervised methods. \textbf{\underline{Bold underline}} denotes the best results, \textbf{bold} denotes the second best results. }
    \label{tab:comparison}
\end{table}

We compare the proposed PaNN with a series of state-of-the-art algorithms, including 1) the fully-supervised approach (denoted as ``-fully-sup''), where we train exclusively only on the fully-labeled dataset $\S_\textup{L}$, 2) the semi-supervised approach  (denoted as ``-semi-sup''), where we train the network on both the fully-labeled dataset $\S_\textup{L}$ and the partially-labeled dataset(s) $\S_\textup{P}$ while treating $\S_\textup{P}$ as unlabeled following the representative method~\cite{bai2017semi}, and 3) the naive partially-supervised approach (denoted as ``-partial-sup''), where we also train the network on both $\S_\textup{L}$ and $\S_\textup{P}$ while treating the partial labels as they are. Different from PaNN, we set $\lambda_2=0$ in Eq.~\eqref{eq:overall_loss} to verify the efficacy of the prior-aware loss. 

\vspace{1ex}\noindent\textbf{Benefit of Partial Supervision.}~As shown from Table~\ref{tab:comparison}, among three kinds of supervisions, partial supervision obtains the best performance followed by the semi-supervision and full supervision. It is no surprise to observe such a phenomenon for two reasons. First, compared with full supervision, semi-supervision has more training data, though part of them is not annotated. Second, compared with semi-supervision, partial supervision involves more annotated pixels in the organ of interest. 

\vspace{1ex}\noindent\textbf{Effect of PaNN.}~From Table~\ref{tab:comparison}, PaNN generally achieves better performance than the naive partially-supervised methods, which demonstrates the effectiveness of our proposed PaNN. For example, when setting the partial dataset as the union of \textbf{A}, \textbf{B} and \textbf{C}, PaNN achieves the best result either using 2D models or 3D models. 2D models generally observe a better performance in each setting compared with 3D models. This is probably due to the fact that current 3D models only act on local patches (\eg, $64 \times 64 \times 64$), which results in lacking holistic information~\cite{wang2018abdominal}. A detailed discussion of 2D and 3D models is listed in~\cite{lai2015deep}. More specifically, PaNN outperforms the naive partially-supervised method by $1.28\%$ with ResNet-50 and by $1.69\%$ with ResNet-101 as the backbone model, respectively. Additionally, we also observe a convincing performance gain of $0.45\%$ using 3D UNet~\cite{cciccek20163d, ronneberger2015u} as the backbone model. 
              
\begin{figure*}
\centering
\includegraphics[width=2\columnwidth]{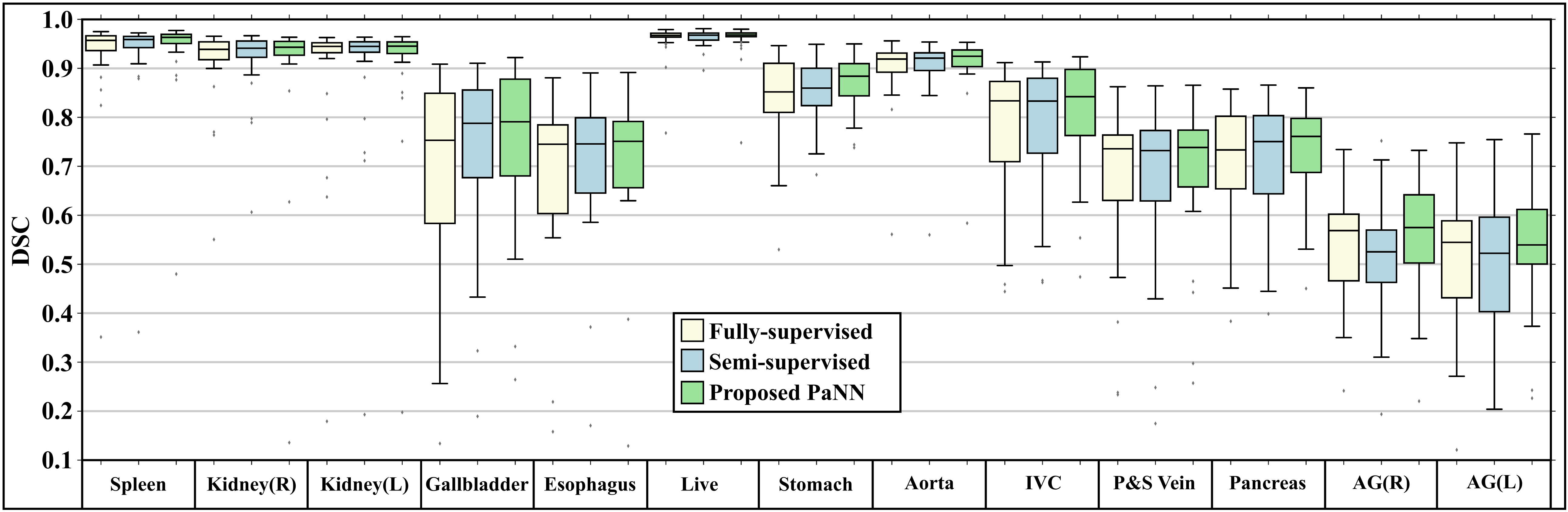}
\caption{Performance comparison (DSC) in box plots of 13 abdominal structures, where the partially-labeled dataset \textbf{C} is used with ResNet-50 as the backbone model. Our proposed PaNN improves the overall mean DSC and also reduces the standard deviation. Kidney/AG (R), Kidney/AG (L) stand for the right and left kidney/adrenal gland, respectively.}
\label{fig:boxplot}
\vspace{-1em}
\end{figure*}

\begin{table*}[!tb]
\small
\centering
\resizebox{\linewidth}{!}{
\begin{tabular}{l|cccccccc|ccc}
\toprule[0.2em]
\multirow{2}{*}{Name}           &\multirow{2}{*}{Spleen}& \multirow{2}{*}{Kidney(R)}  &\multirow{2}{*}{Kidney(L)} &\multirow{2}{*}{Gallbladder} &\multirow{2}{*}{Esophagus} &\multirow{2}{*}{Liver}  &\multirow{2}{*}{Aorta} &\multirow{2}{*}{IVC} &Average & Mean Surface & Hausdorff \\
         & &  & && & & & &Dice    & Distance & Distance \\
\midrule
AutoContext3DFCN~\cite{roth2018multi} &0.926 &0.866  &0.897  &0.629 & 0.727 &
0.948  & 0.852  &0.791                                            & 0.782  & 1.936 & 26.095            \\ 
deedsJointCL~\cite{heinrich2015multi}       &0.920  &0.894  &0.915  &0.604  &0.692  &0.948 &
0.857    &0.828                                                    & 0.790 & 2.262  & 25.504            \\
dltk0.1\_unet\_sub2~\cite{pawlowski2017dltk}      &0.939 &0.895 &0.915 & \textbf{0.711}  &0.743    &0.962   &0.891 &0.826	                                                & 0.815 & 1.861 & 62.872            \\
results\_13organs\_p0.7         &0.890 &  0.898 &0.883  &0.685   &0.754 &0.936   &0.870    &0.819                                                        & 0.817 & 4.559  & 38.661            \\
PaNN* (\textbf{ours})   & \textbf{0.961}  & \textbf{0.901} & \textbf{0.943} & 0.704  & \textbf{0.783}    & \textbf{0.972}      & \textbf{0.913}   & \textbf{0.835}                                                                & \textbf{0.832} & \textbf{1.641} & \textbf{25.176}             \\
PaNN (\textbf{ours})     &\underline{\textbf{0.968}} & \underline{\textbf{0.920}} & \underline{\textbf{0.953}}  & \underline{\textbf{0.729}}   & \underline{\textbf{0.790}}      & \underline{\textbf{0.974}}   & \underline{\textbf{0.925}}  & \underline{\textbf{0.847}}                                                                   & \underline{\textbf{0.850}} & \underline{\textbf{1.450}} & \underline{\textbf{18.468}}             \\
\bottomrule[0.2em]
\end{tabular}
}
\caption{Performance comparison on the 2015 MICCAI Multi-Atlas Abdomen Labeling challenge leaderboard. Our method achieves the largest Dice score and the smallest average surface distances and Hausdorff distances. PaNN* only uses $80\%$ of the training data
as the fully-supervised dataset and use the rest $20\%$ data
as partially-labeled data (by randomly removing labels of
8/13 organs), without using extra data. In this table, we only show 8/13 organs' average Dice scores due to the space limit.}
\label{tbl:comparison}
\vspace{-1em}
\end{table*}

Meanwhile, by increasing the number of partially-labeled datasets (from using only \textbf{A}, \textbf{B} or \textbf{C} to the union of three), the performance improvements of different methods are also different. For example, with the ResNet-101 as the backbone, the largest improvement obtained under semi-supervision is $0.82\%$ (from $76.37\%$ to $77.19\%$), and that of partial supervision is $0.51\%$ (from $76.84\%$ to $77.35\%$). By contrast, PaNN obtains a much more remarkable improvement of $1.56\%$ (from $77.48\%$ to $79.04\%$). Such an observation suggests that PaNN is capable of handling more partially-labeled training data and is less susceptible to the background ambiguity.

\vspace{1ex}\noindent\textbf{Organ-by-organ Analysis.}~To reveal the detailed effect of PaNN, we present an organ-by-organ analysis in Fig.~\ref{fig:boxplot}. We use ResNet-50 as the backbone model (ResNet-101 has a similar trend) and the partially-labeled dataset~\textbf{C} (indicates that the liver is the target organ).

In Fig.~\ref{fig:boxplot}, we observe clear statistical improvements over the fully-supervised method for almost every organ (p-values $ p < 0.001$ hold for 11/13 of all abdominal organs). Great improvements are also observed for those difficult organs,~\ie, organs either in small sizes or with complex geometric characteristics such as gallbladder (from $67.26\%$ to $72.26\%$), esophagus (from $69.35\%$ to $71.21\%$), stomach (from $84.09\%$ to $87.21\%$), IVC (from $77.34\%$ to $80.70\%$), portal vein \& splenic vein (from $66.74\%$ to $68.75\%$), pancreas (from $71.45\%$ to $73.62\%$), right adrenal gland (from $53.65\%$ to $55.56\%$) and left adrenal gland (from $49.51\%$ to $53.63\%$). This promising result indicates that our method distills a reasonable amount of knowledge from additional partially-labeled data and the regularization loss can help facilitate the network to enhance the discriminative information to a certain degree.

Meanwhile, we also observe a distinct performance improvement for organs other than the partially-labeled structures (\ie, the liver). For instance, the performance of gallbladder, stomach, IVC, pancreas are boosted from $68.97\%$, $85.57\%$, $78.59\%$, $71.94\%$ to $72.26\%$, $87.21\%$, $80.70\%$, $73.62\%$, respectively. This suggests that the superiority of PaNN not only originates from more training data, but also from the fact that PaNN can effectively incorporate anatomical priors on organ sizes in abdominal regions, which is helpful for multi-organ segmentation.

\begin{figure*}[tb]
    \vspace{-1em}
 	\centering
 	\includegraphics[width=1.86\columnwidth]{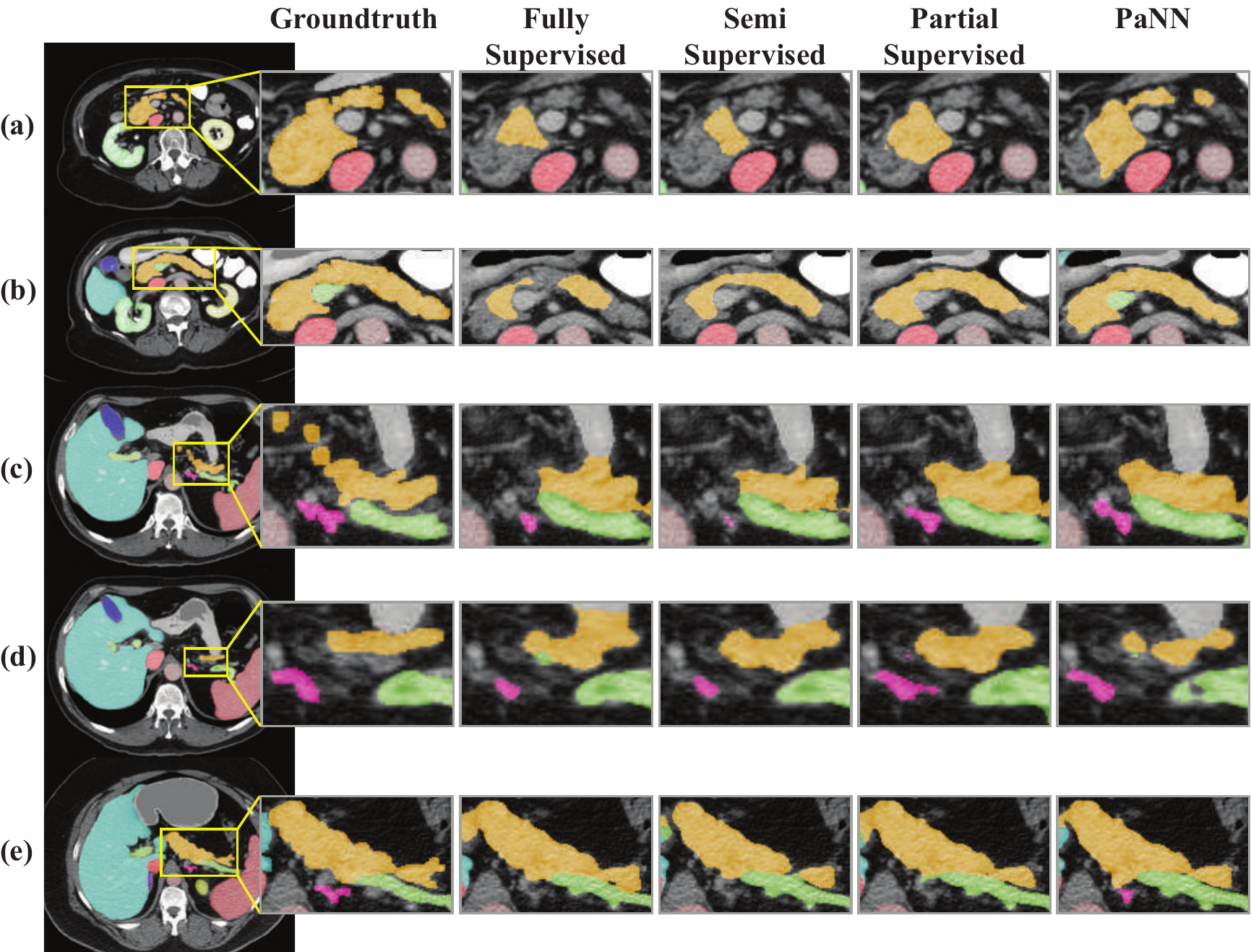}
 	\caption{Qualitative comparison of different methods, where the partially-labeled dataset \textbf{C} is used as partial supervision with ResNet-101 as the backbone model. We exhibit 3 cases (5 slices) as examples. Improved segmentation regions are zoomed in from the axial view to demonstrate finer details. 
}
 	\label{fig:qualitative}
 	\vspace{-1em}
\end{figure*}

\vspace{1ex}\noindent\textbf{Qualitative Evaluation.}~We also show a set of qualitative examples,~\ie,~$5$ slices from $3$ cases, in Fig.~\ref{fig:qualitative}, where we zoom in to visualize the finer details of the improved region. 

In these samples, we observe that PaNN is the only method that successfully detects the pancreatic tail in Fig.~\ref{fig:qualitative}(a). In Fig.~\ref{fig:qualitative}(b), all other methods fail to detect the portal vein and splenic vein while PaNN demonstrates an almost perfect detection of these veins. For Fig.~\ref{fig:qualitative}(c) to Fig.~\ref{fig:qualitative}(e), apart from the evident improvements of the pancreas, left adrenal gland, one of the smallest abdominal organs, is also clearly segmented by PaNN.

\subsection{MICCAI 2015 Multi-Atlas Labeling Challenge}
We test our model in the 2015 MICCAI Multi-Atlas Abdomen Labeling challenge. The top model (denoted as ``PaNN'' in Table \ref{tbl:comparison}) we submit is based on ResNet-101, and trained on all 30 cases of the fully-labeled dataset $\S_L$ and the union of three partially-labeled datasets \textbf{A}, \textbf{B} and \textbf{C}. The evaluation metric employed in this challenge includes the Dice scores, average surface distances~\cite{roth2018spatial} and Hausdorff distances~\cite{milletari2016v}. We compare PaNN with the other top submissions of the challenge leaderboard in Table~\ref{tbl:comparison}. As it shows, the proposed PaNN achieves the best performance under all the three evaluation metrics, easily surpassing prior best result by a large margin. \textbf{Without using any additional data and even randomly removing partial labels} from the challenge data, our method (denoted as ``PaNN*'' in Table \ref{tbl:comparison}) stills obtains the state-of-the-art result of ~$83.17\%$, outperforming the previous best result of DLTK UNet~\cite{pawlowski2017dltk} by ~$2\%$ in average Dice. It is noteworthy that our method is far from its potential maximum performance as we only use 2D single view algorithms. It is suggested~\cite{zhou2017fixed, wang2018abdominal,zhou2019semi} that using multi-view algorithms or model ensemble can boost the performance further.

\subsection{Generalization to Other Datasets}
\label{sec:generalization}

\begin{table}[!tb]
\small
\centering
\resizebox{\linewidth}{!}{
\begin{tabular}{lcccc}
\toprule[0.2em]
 \multirow{3}{*}{Organ}      & Fully    & Semi  & Partially & PaNN  \\
                             & Supervised    & Supervised  & Supervised (\textbf{ours}) & (\textbf{ours}) \\
\midrule
Gallbladder         &0.8225 & 0.8399      &0.8465         & \underline{\textbf{0.8467} }           \\
Aorta                &0.9110  & 0.9096    &0.9121        & \underline{\textbf{0.9133}}             \\
IVC                  & 0.8083 & 0.8175    &0.7995       & \underline{\textbf{0.8266}}          \\
Pancreas              &0.7831 & 0.7994    &0.8079         & \underline{\textbf{0.8193}}            \\
\midrule
avg. Dice              &0.9008 & 0.9060    &0.9063         & \underline{\textbf{0.9103}}            \\
\toprule[0.2em]
\end{tabular}
}
\caption{Performance comparison on a newly collected dataset. Full results are included in the appendix.}
\label{tbl:generalization}
\vspace{-1.7em}
\end{table}

We also apply our algorithm to a different set of abdominal clinical CT images, where 20 cases are used for training and 15 cases are used for testing. A total of 9 structures (spleen, right kidney, left kidney, gallbladder, liver, stomach, aorta, IVC, pancreas) are manually labeled. Each case was segmented by four experienced radiologists, and confirmed by an independent senior expert. Each CT volume consists of $319 \sim 1051$ slices of $512 \times 512$ pixels, and has voxel spatial resolution of $([0.523 \sim 0.977] \times [0.523 \sim 0.977]\times 0.5)mm^3$. We use the union of all 3 datasets \textbf{A}, \textbf{B}, and \textbf{C} as the partial supervision. The results are summarized in Table~\ref{tbl:generalization}, where the proposed PaNN also achieves better results compared with existing methods.

\section{Conclusion} \label{sec:con}
In this work, we have presented PaNN, for multi-organ segmentation, as a way to better utilize existing partially-labeled datasets. In several applications such as radiation therapy or computer-aided surgery, physicians and surgeons have been doing segmentation of target structures. Meanwhile, to handle the background ambiguity brought by the partially-labeled data, the proposed PaNN exploits the anatomical priors by regularizing the organ size distributions of the network output should approximate their prior statistics in the abdominal region. Our proposed PaNN shows promising results using state-of-the-art models. 

\noindent {\bf Acknowledgements.}
This work was partly supported by the Lustgarten Foundation for Pancreatic Cancer Research.

{\small
\bibliographystyle{ieee}
\bibliography{egbib}
}

\clearpage
\onecolumn
\appendix

\section*{A. Summary of Partially-labeled Datasets}
\label{sec:list}
To facilitate the research on partially-supervised multi-organ segmentation, we have collected a list of partially-labeled datasets to the best of our knowledge. We present them in Table~\ref{tbl:comparison} for the fellow researchers to explore partial supervision in multi-organ segmentation problems by leveraging these partially-labeled datasets. Note that our method can be also easily applied to these datasets for improving the segmentation performance.
\begin{table}[h]
\small
\centering
\resizebox{\columnwidth}{!}{
\tiny
\begin{tabular}{|l|l|c|p{6.8cm}|}
\hline
Name & Target Organs   & N & Link \\
\hline
\hline
Anatomy3 & \makecell[l]{ Liver\\Pancrea\\Stomach\\Spleen\\Gallbladder} & 20 & \url{http://www.visceral.eu/benchmarks/anatomy3-open/}\\
\hline
Pancreas-CT & Pancreas & 82 & \url{https://wiki.cancerimagingarchive.net/display/Public/Pancreas-CT} \\
\hline
ISBI 2019 Challenge  & Liver & 30 & \url{https://chaos.grand-challenge.org/Data/} \\
\hline
\multirow{3}{*}{\makecell[l]{Medical Segmentation \\Decathlon}} & Liver & 131 &  \multirow{3}{*}{\url{http://medicaldecathlon.com}}\\
                                & Spleen & 41 &  \\
                                & Pancreas & 282 &  \\
\hline
Sliver07    & Liver & 20 & \url{http://sliver07.org/} \\
\hline
MICCAI 2017 LiTS    & Liver & 131 &  \url{https://competitions.codalab.org/competitions/17094} \\
\hline
\end{tabular}
}
\caption{Summary of partially-labeled datasets for multi-organ segmentation. N denotes the number of annotated cases.}
\label{tbl:comparison}
\end{table}


\section* {B. Qualitative Evaluation}
\label{sec:visualization}
\begin{figure*}[tb]
 	\centering
 	\includegraphics[width=0.7\linewidth]{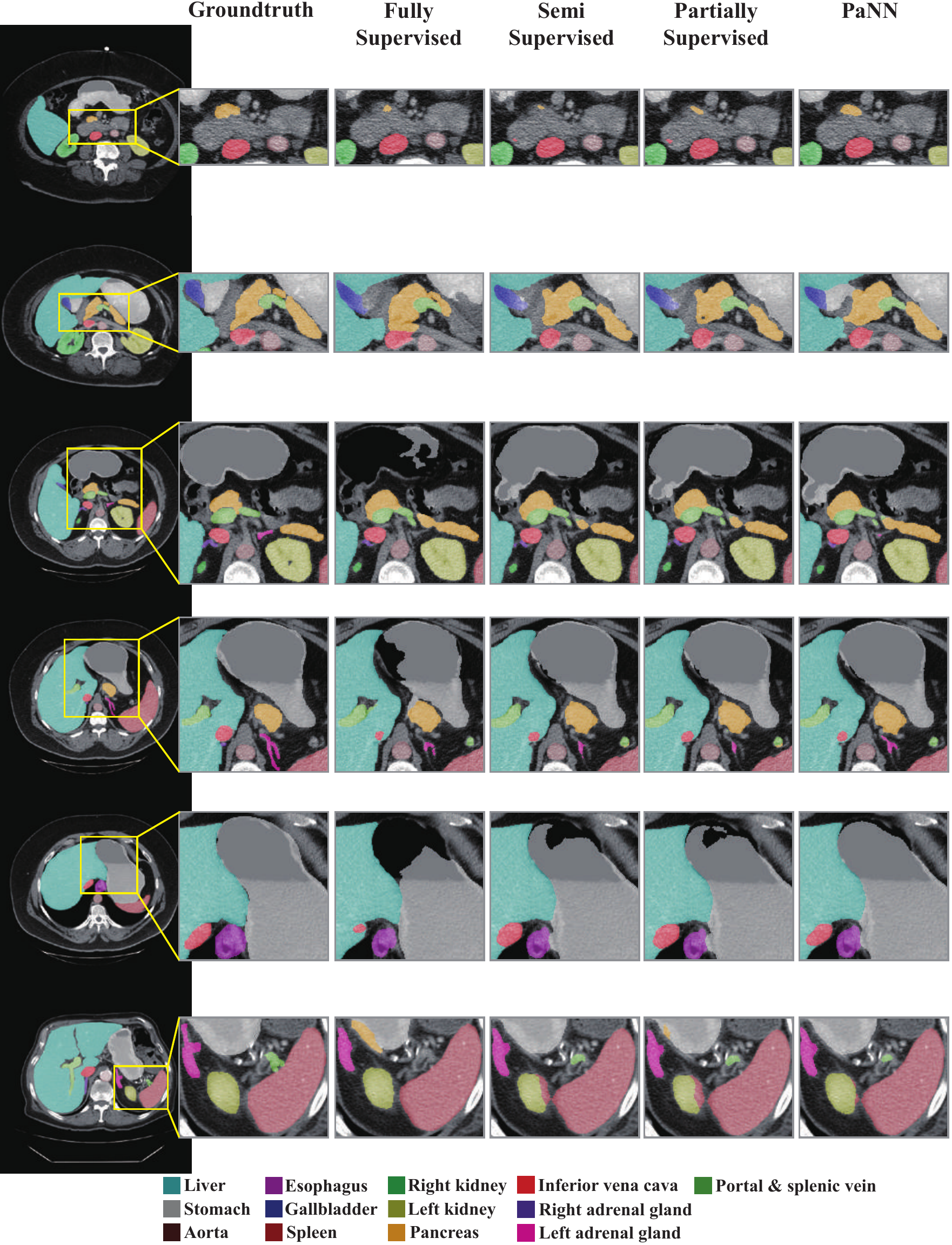}
 	\caption{Qualitative comparison of different methods, where all the 3 partially-labeled datasets \textbf{A},\textbf{B},\textbf{C} are used as the partial supervision with ResNet-101 as the backbone model. We exhibit 5 cases (6 slices) as examples. Improved segmentation regions are zoomed in from the axial view to demonstrate finer details. Besides the pancreas, portal vein \& splenic vein, left adrenal gland, we also show other improved regions such as the stomach, gallbladder,~\etc.}
 	\label{fig:qualitative_suppl}
 	\vspace{-1em}
\end{figure*}

We include more qualitative results in Fig.~\ref{fig:qualitative_suppl}, where we also show improved regions compared with fully-supervision other than the pancreas, portal vein \& splenic vein, left adrenal gland as discussed in the main manuscript. In Fig.~\ref{fig:qualitative_suppl}, we can see an evident improvement of the pancreas (row 1-3), the gallbladder (row 2), the left adrenal gland (row 3-4), the stomach (row 3-6), and the portal vein \& the splenic vein (row 6).

\section*{C. Generalization to Other Datasets}
The \textbf{complete} results of Table~3 in the main manuscript are summarized in Table~\ref{tbl:generalization}, where the proposed PaNN also achieves the best performance compared with existing methods.

\begin{table}[!hb]
\small
\centering
\resizebox{0.55\linewidth}{!}{
\tiny
\begin{tabular}{lcccc}
\toprule[0.2em]
 \multirow{3}{*}{Organ}      & Fully    & Semi  & Partially & PaNN  \\
                             & Supervised    & Supervised  & Supervised (\textbf{ours}) & (\textbf{ours}) \\
\midrule
Spleen               &0.9640  &0.9651     &0.9673         & 0.9666           \\ 
Right kidney         &0.9626  &0.9627     &0.9625         & 0.9615            \\
Left kidney          &0.9530  & 0.9547    &0.9526         & 0.9541            \\
Gallbladder         &0.8225 & 0.8399      &0.8465         & 0.8467            \\
Liver                &0.9684 & 0.9691  &0.9691 & 0.9689            \\
Stomach             &0.9344            & 0.9363   &0.9396 & 0.9361            \\
Aorta                &0.9110  & 0.9096    &0.9121        & 0.9133             \\
IVC                  & 0.8083 & 0.8175    &0.7995       & 0.8266           \\
Pancreas              &0.7831 & 0.7994    &0.8079         & 0.8193            \\
\midrule
avg. Dice              &0.9008 & 0.9060    &0.9063         & \textbf{0.9103}            \\
\toprule[0.2em]
\end{tabular}
}
\caption{Performance comparison on a newly collected high-quality abdominal dataset, where our method achieves the best result.
}
\label{tbl:generalization}
\vspace{-1.6em}
\end{table}

\end{document}